# CheXPoT: Post-Hoc Visual Verification and Correction of VLM-Generated Radiology Reports

Joey Guo[1], Rayan Ansari[2], Alaa Youssef[3]

[1]The Nueva School [2]Stanford Cardiovascular Institute [3]Department of Radiology, Stanford School of Medicine

**Abstract**

Vision-language models (VLMs) have shown promise for chest radiograph report generation, but adoption remains limited because of hallucinated findings that are not grounded in the image. Correcting these hallucinations and improving overall report quality remains a significant challenge. Prior work has addressed this problem by refining models with reinforcement learning, preference optimization, or chain-of-thought supervision, often at substantial compute cost. To address this gap, we introduce CheXPoT, a program-of-thought (PoT) framework that verifies each generated finding against image features and removes those it cannot ground. CheXPoT decomposes generated reports into predicates with RadGraph and an LLM, grounds these predicates against a DenseNet-121 and Grad-CAM visual evidence table, executes PoT verification code, and rewrites reports after unsupported predicates are removed. On a 4,145-image held-out split, CheXPoT yields statistically significant precision gains on 8 of 14 CheXpert classes and F1 gains on 5, with no significant decreases in either metric. CheXPoT lifts micro-average precision from 19.2% (MedGemma-4B) to 46.8% (MedGemma-4B + CheXPoT) and micro-average F1 from 23.0% to 49.4%, absolute gains of 27.6 and 26.4 percentage points, respectively. Aside from MedGemma, we also tested CheXPoT on other report generators. On LLaVA-Med, CheXPoT lifts micro-average precision from 4.5% to 17.7% and micro-average F1 from 6.3% to 19.4%, absolute gains of 13.2 and 13.1 percentage points, respectively. On MAIRA-2, CheXPoT lifts micro-average precision from 15.6% to 42.1% and micro-average F1 from 21.0% to 45.3%, absolute gains of 26.5 and 24.3 percentage points, respectively. Our results suggest that inexpensive post-hoc visual verification may improve VLM trustworthiness by correcting unsupported findings from generated radiology reports. CheXPoT lays a foundation for safer and more trustworthy tools that help validate VLM-generated radiology reports, reduce unsupported findings, and support radiologists in model-assisted workflows. These innovations may subsequently help increase trust and adoption of AI VLM tools in Medicine.

## 1 Introduction

Vision-language models (VLMs) have shown promise for chest X-ray report generation, but they have not yet proven sufficient for routine clinical use. A generated report may appear complete while still containing unsupported findings that do not match the image. For clinicians, completeness is therefore inseparable from trust: a report is difficult to use if its findings are not visually grounded.

This work asks a narrower question: can an existing radiology VLM report be made more trustworthy after generation, without retraining the full generator? We introduce **CheXPoT**, a post-hoc verification framework for AI-generated radiology reports on CheXpert Plus that checks generated findings against image-derived evidence and rewrites the report after removing unsupported claims.

The motivating failure mode is hallucination. Radiology VLMs can produce fluent findings that are not present in the chest X-ray, creating a risk that an anatomically plausible report overstates the evidence in the image [1–3]. Continuous hallucination scores used in prior work can be useful for ranking, but they do not directly produce a corrected report for downstream use. [1,9]

Prior work has approached hallucination through structured generation, reasoning supervision, reinforcement learning, direct preference optimization, latent steering, and post-hoc hallucination detection [1, 4–16]. ReXTrust, for example, provides fine-grained hallucination detection for individual report findings [9]. Our work is complementary: rather than only scoring suspect findings, CheXPoT uses visual verification to revise the report. This design is intended to be lightweight, modular, and compatible with existing report generators.

Recently, methods such as Pelican, utilizing Program-of-Thought prompting and computer vision models, have shown to be able to reduce hallucinations at general tasks, but have not yet been applied for Chest X-Rays [17]. We developed CheXPoT, a specialized LLM and CNN-based framework to decrease hallucinations and enhance reports. Our contributions are:

1. We introduce **CheXPoT**, a PoT-based verification framework that decomposes reports into predicates with RadGraph [18], grounds them using DenseNet-121 [19] and GradCAM [20] visual evidence, and rewrites reports after unsupported predicates are removed.

2. On a 4,145-image held-out split, CheXPoT significantly improves precision for 8 of 14 CheXpert labels and F1 for 5 of 14 labels, with no statistically significant decreases in precision or F1. We show that CheXPoT + MedGemma improves micro-average precision from 19.2% to 46.8%, **a net gain of 27.6 percentage points**, and micro-average F1 from 23.0% to 49.4%, **a net gain of 26.4 percentage points**, relative to MedGemma-4B generated reports.

3. We also extended our approach to run on multiple models. For LLaVA-Med, we show that CheXPoT improves micro-average precision from 4.5% to 17.7%, a net gain of 13.2 percentage points, and micro-average F1 from 6.3% to 19.4%, a net gain of 13.1 percentage points relative to LLaVA-Med. For MAIRA-2, we show that CheXPoT improves micro-average precision from 15.6% to 42.1%, a net gain of 26.5 percentage points, and micro-average F1 from 21.0% to 45.3%, a net gain of 24.3 percentage points relative to MAIRA-2.

# 2 Methods

## 2.1 Data Description

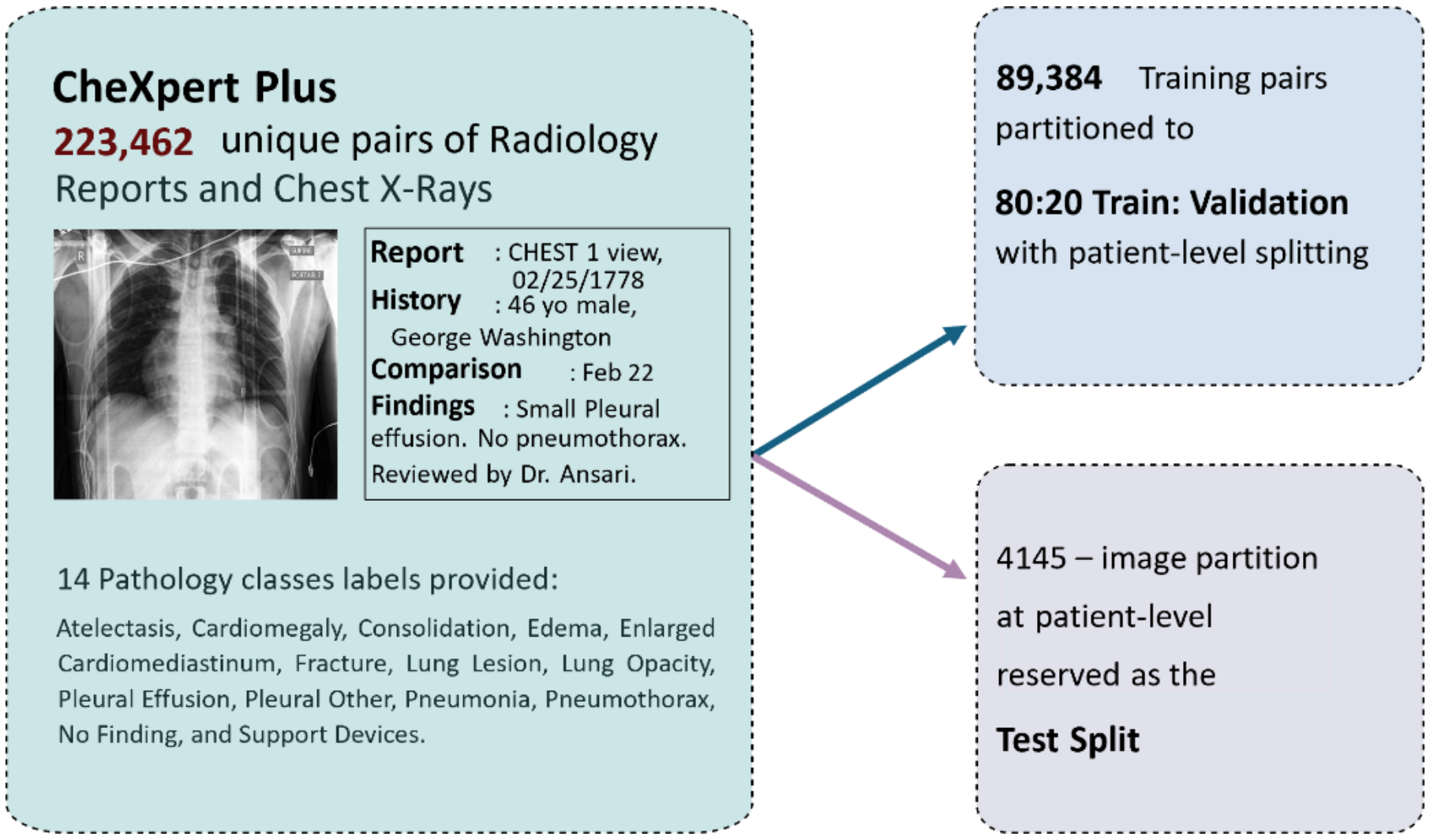

*Figure 1: Data preprocessing for CheXPoT. CheXpert Plus image–report pairs are prepared for visual verification: CheXpert Plus labels support DenseNet-121 training.*

CheXPoT uses CheXpert Plus image–report pairs to train and evaluate the visual verification pipeline. CheXpert Plus is a publicly available open dataset that contains 223,462 unique pairs of radiology reports and chest X-rays across 187,711 studies from 64,725 patients, providing the image–report pairs needed for visual verification [21,22]. CheXpert Plus also provides three sets (Impression, Finding, Report) of Pathology labels for the 14 classes: Atelectasis, Cardiomegaly, Consolidation, Edema,

Enlarged Cardiomediastinum, Fracture, Lung Lesion, Lung Opacity, Pleural Effusion, Pleural Other, Pneumonia, Pneumothorax, No Finding, and Support Devices. These labels were obtained using CheXbert on the radiologist reports on the specified section. Impression labels were generated only on the Impression section of the report, Finding labels were generated only on the Finding section of the report, and Report labels were generated on the entire report.

We elected to choose CheXpert Plus for our dataset for the following reasons:

1. CheXpert Plus has 223,462 image-report pairs, which is significantly more data than other datasets.
2. CheXpert Plus provides both X-Rays and Radiologist Reports, which is necessary for the Vision Language Modeling task that we worked on.
3. CheXpert Plus provides pre-generated Pathology labels at a high accuracy, which is necessary to be able to train a DenseNet-121 model.

We restrict this study to CheXpert Plus so that verification and report-level evaluation are performed within a consistent dataset and label ecosystem.

## 2.2 CheXPoT Framework

**Claim decomposition.** The first step of the CheXPoT pipeline is to decompose the report into predicates. To accomplish this task, the report is first divided into entities by RadGraph [18], and then the list of entities is converted to a list of predicates by an LLM (gpt-4o-mini).

**Visual table construction.** The second step of the CheXPoT pipeline is to identify key pathologies in the chest radiograph. We utilize a fine-tuned DenseNet-121 model [19], finetuned on our designated train split of CheXpert Plus, to predict probabilities for the 14 CheXpert labels. Using the thresholds described in section 2.3, the probabilities are turned into class presences. Then, using Grad-CAM [20], the most representative points for each class are extracted. Pathology name, presence, and representative point are compiled into a pandas dataframe.

**Program-of-Thought verification.** The third step of the CheXPoT pipeline is to obtain answers to the predicates obtained from the claim decomposition step utilizing the visual table. To do this, we utilize Program-of-Thought prompting. We prompt the LLM (gpt-4.1) to generate code based on the predicates and the structure of the visual table, but not the contents. After the code has been generated, the contents of the visual table are passed through to obtain the answers to the questions. Inconclusive answers are discarded.

**Report correction.** The fourth step of the CheXPoT pipeline is to corroborate the original report against the new answers to the predicates. Both are passed to an LLM (gpt-4o-mini) that is instructed to correct the report or leave it as is if the information matches. The corrected report is then outputted.

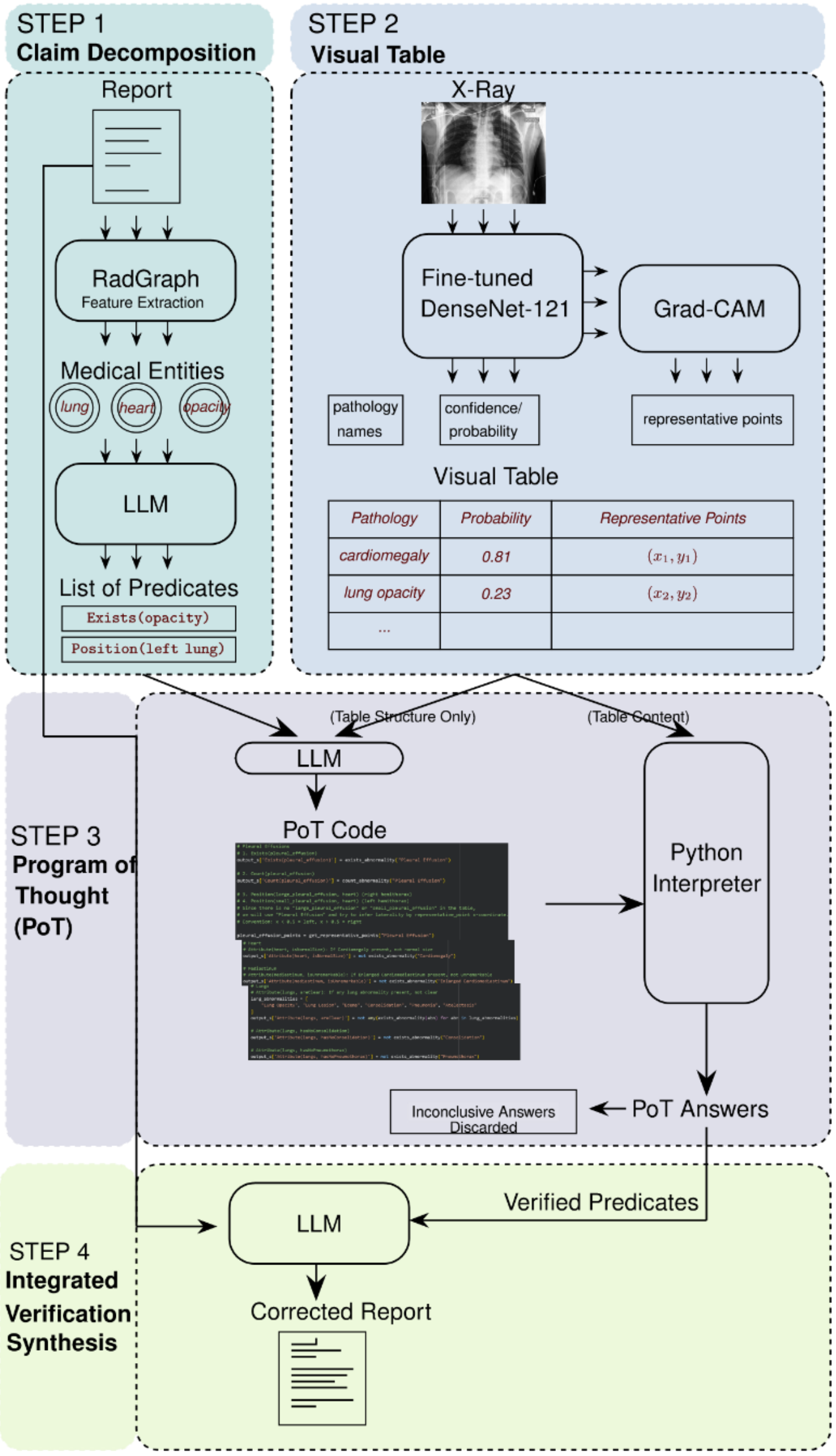

*Figure 2: CheXPoT Architecture.*

## 2.3 Training

**DenseNet fine-tuning.** To be able to generate visual tables, we fine-tuned a DenseNet-121 model on the 14 CheXpert pathologies. Prior to selecting the DenseNet-121 model, we also tried ResNet-18, ResNet-50, ViT, and MedImageInsight models. We finally selected the DenseNet-121 model as it achieved a significantly higher macro-F1 score on the validation set. The model's final fully connected layer was modified to have 14 output channels. The training set, CheXpert Plus Chunks 0 and 2, was split into an 80/20 train-validation split. For ground-truth labels, we utilized the Finding labels set as it is the most accurate for the image classification task as purported by the authors of CheXpert Plus. [21,22]

**Optimization.** The DenseNet-121 model was trained for 32 epochs with a batch size of 32, using focal loss [24] with $\alpha = 1$ and $\gamma = 2$, the AdamW optimizer [25] with $10^{-5}$ weight decay, and a ReduceLROnPlateau learning rate scheduler with base learning rate $10^{-5}$. We addressed leakage by ensuring strict patient-level splits. We used focal loss to counter the large class imbalances (e.g. the smallest class, Pleural Other, consists of only 1.58% of the entire dataset)

**Thresholds.** To generate classification thresholds, we computationally found the thresholds that yielded the highest per-class F1s on the validation split.

## 2.4 Evaluation

**Report-level comparison.** To compare the performance of CheXPoT-corrected VLM reports with baseline VLM-generated reports, we first ran MedGemma-4B [26] zero-shot on our designated test split, containing 4,145 images. Afterwards, we ran the CheXPoT pipeline on the generated MedGemma reports and images, producing the corrected reports.

Running the CheXPoT Pipeline on all 4145 reports took 618 minutes and $26.77 in OpenAI API Spend. Averaging over every individual report, the average study took 8.95 seconds and $0.00646 to run.

For Ground-Truth labels, we utilized the already generated labels attached to the CheXpert Plus Dataset, specifically the Findings Label set. To label the newly generated reports, we used CheXbert.

We estimated uncertainty using a paired nonparametric bootstrap. In each of 10,000 iterations, studies were sampled with replacement, and the MedGemma-4B and CheXPoT-corrected reports were evaluated on the same resampled cases. We calculated the paired difference in precision, recall, and F1 between the two systems. The 2.5th and 97.5th percentiles of the resulting difference distribution defined the 95% confidence interval. A difference was considered statistically significant when its paired 95% bootstrap confidence interval excluded zero. An interval with an endpoint equal to zero was treated as non-significant.

# 3 Results

Table 1 summarizes the per-label CheXbert evaluation for CheXPoT and the MedGemma-4B baseline before presenting the bootstrap confidence interval plots. CheXPoT improves the micro-average precision, recall, and F1, with the largest aggregate gain in F1. Take Lung Opacity as an example to show the effectiveness of the CheXPoT framework. Lung Opacity precision was 15.7% with MedGemma alone, but after CheXPoT report correction, Lung Opacity precision is 35.8%, **a gain of 20.1%.** Looking at F1 score, Lung Opacity F1 was 23.0% with MedGemma alone, but after CheXPoT report correction, Lung Opacity F1 is 34.8%, **a gain of 11.8%.** Other pathology labels show even larger gains.

| Label | Support | CheXPoT + MedGemma-4B | | | MedGemma-4B | | |
|---|---|---|---|---|---|---|---|
| | | Prec. | Rec. | F1 | Prec. | Rec. | F1 |
| Enlarged Cardiomediastinum | 71 | **0.030** | **0.070*** | **0.042** | 0.029 | 0.056 | 0.038 |
| Cardiomegaly | 188 | **0.232*** | 0.351 | **0.280*** | 0.122 | 0.532 | 0.198 |
| Lung Opacity | 578 | **0.358*** | 0.339 | **0.348*** | 0.157 | 0.433 | 0.230 |
| Lung Lesion | 50 | 0.000 | 0.000 | 0.000 | 0.029 | 0.020 | 0.024 |
| Edema | 241 | **0.330*** | 0.120 | **0.176** | 0.113 | 0.195 | 0.143 |
| Consolidation | 81 | **0.067** | 0.025 | 0.036 | 0.039 | 0.049 | 0.044 |
| Pneumonia | 13 | **0.065*** | 0.154 | **0.091*** | 0.011 | 0.154 | 0.021 |
| Atelectasis | 172 | 0.044 | 0.058 | 0.050 | 0.044 | 0.116 | 0.064 |
| Pneumothorax | 101 | **0.349*** | 0.149 | **0.208** | 0.085 | 0.317 | 0.134 |
| Pleural Effusion | 444 | **0.376*** | 0.223 | **0.280** | 0.176 | 0.441 | 0.252 |
| Pleural Other | 17 | 0.000 | 0.000 | 0.000 | 0.000 | 0.000 | 0.000 |
| Fracture | 56 | 0.000 | 0.000 | 0.000 | 0.000 | 0.000 | 0.000 |
| Support Devices | 630 | **0.206*** | 0.767 | **0.325*** | 0.191 | 0.867 | 0.314 |
| No Finding | 3210 | **0.860*** | **0.672*** | **0.755*** | 0.816 | 0.145 | 0.247 |
| Micro avg. | 5852 | **0.468** | **0.524** | **0.494** +0.264 | 0.192 | 0.285 | 0.230 |
| Macro avg. | 5852 | **0.208** | 0.209 | **0.185** +0.063 | 0.129 | 0.238 | 0.122 |
| Weighted avg. | 5852 | **0.588** | **0.524** | **0.527** +0.298 | 0.510 | 0.285 | 0.229 |
| Samples avg. | 5852 | **0.472** | **0.595** | **0.500** +0.304 | 0.210 | 0.211 | 0.196 |

*Table 1: Per-label CheXbert evaluation of MedGemma-4B reports before and after CheXPoT correction on the 4,145-image held-out test split. Green bold values indicate metrics where CheXPoT improves over the MedGemma-4B baseline. Support is identical for both systems and denotes the number of positive label instances in the reference labels.*

We used a bootstrap test, sampling from the test dataset with replacement 10,000 times and calculating the difference between CheXPoT's and MedGemma's metrics. Fig. 3 presents CheXPoT's 95% confidence intervals and average performance. Figs. 4–6 illustrate the changes in precision, recall, and F1 on MedGemma reports after applying CheXPoT.

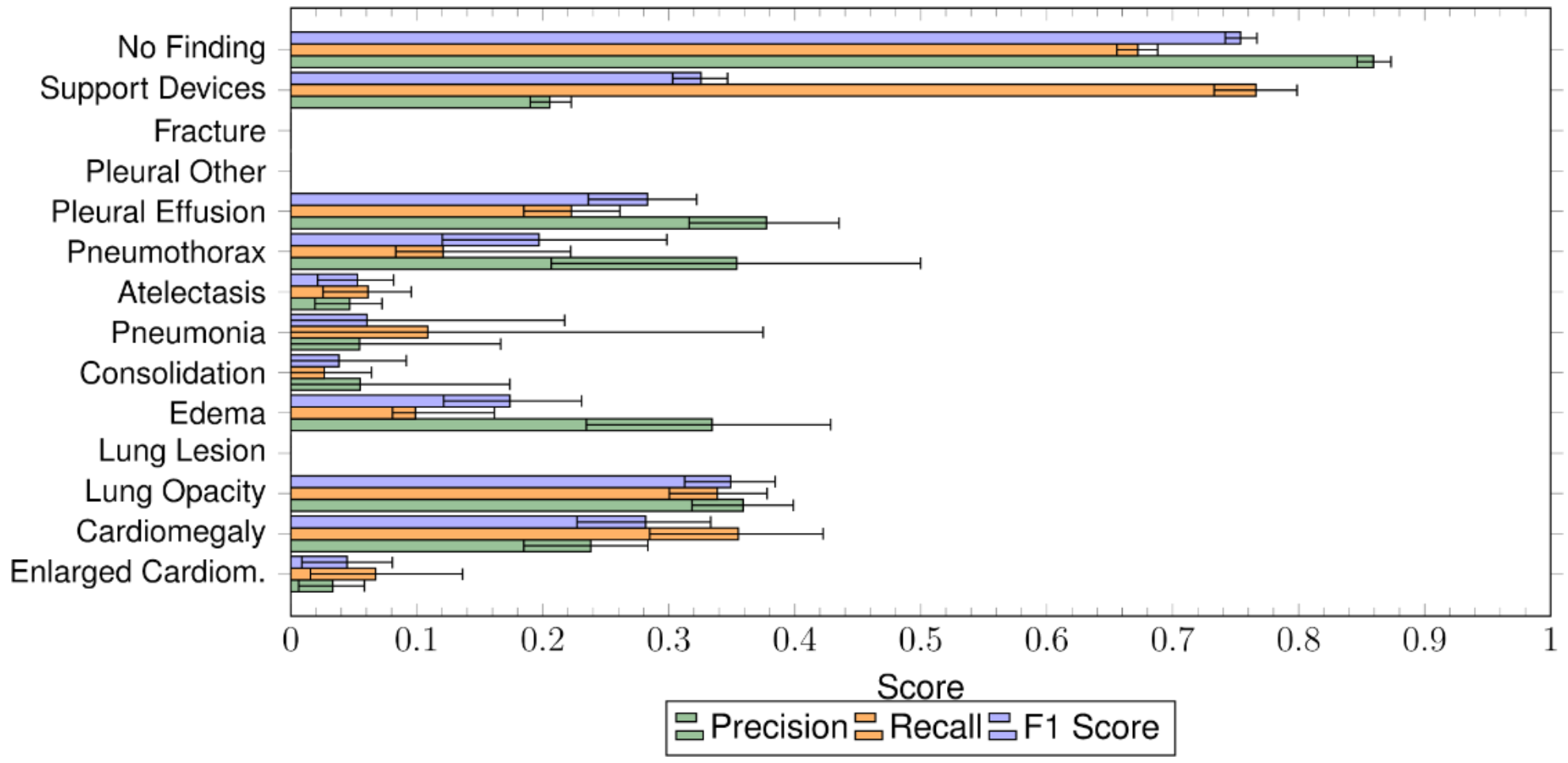

*Figure 3: 95% confidence intervals for Precision, Recall, and F1 Scores of CheXPoT on the 14 CheXpert classes.*

CheXPoT significantly improved precision for eight CheXpert classes: Cardiomegaly, Lung Opacity, Edema, Pneumonia, Pneumothorax, Pleural Effusion, Support Devices, and No Finding. A subset of these gains also translated into significantly improved F1 for five classes: Cardiomegaly, Lung Opacity, Pneumonia, Support Devices, and No Finding. No statistically significant decreases were observed for precision or F1 for all classes.

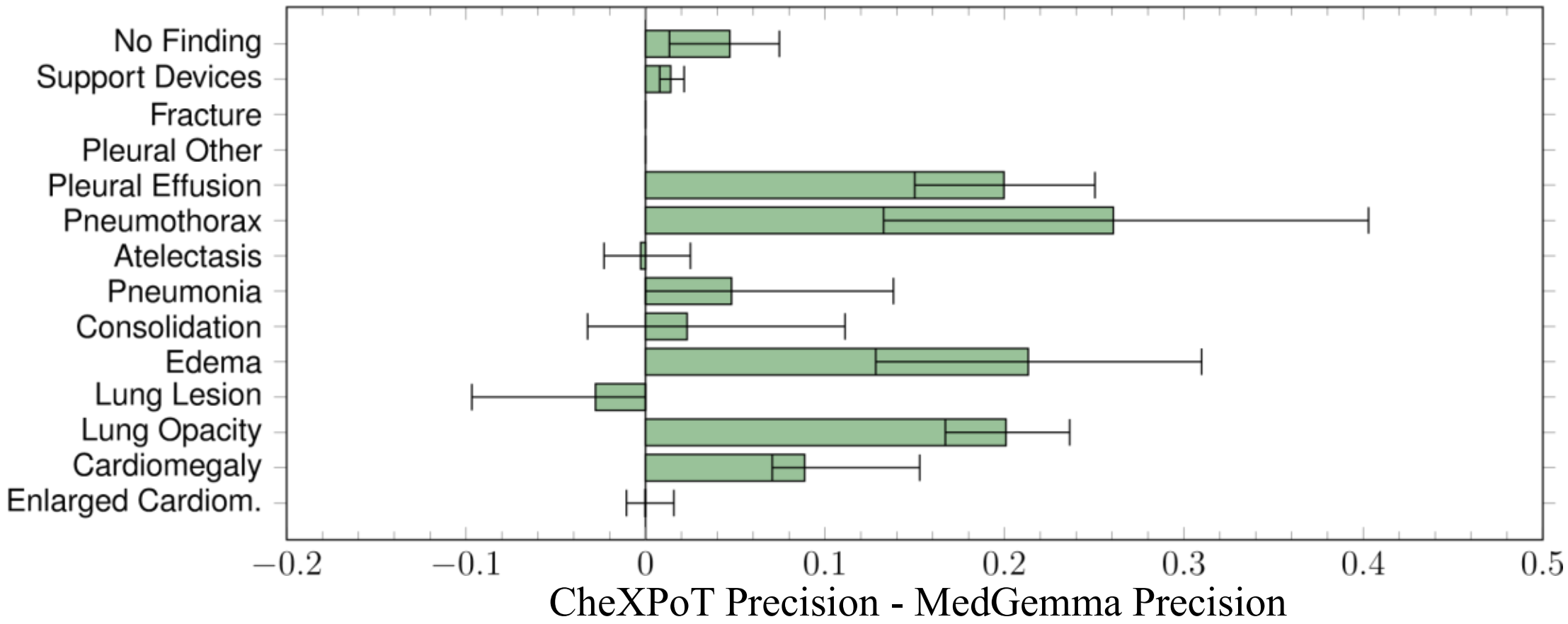

*Figure 4: 95% confidence intervals for difference in Precision between CheXPoT corrected reports and MedGemma Reports on the 14 CheXpert classes.*

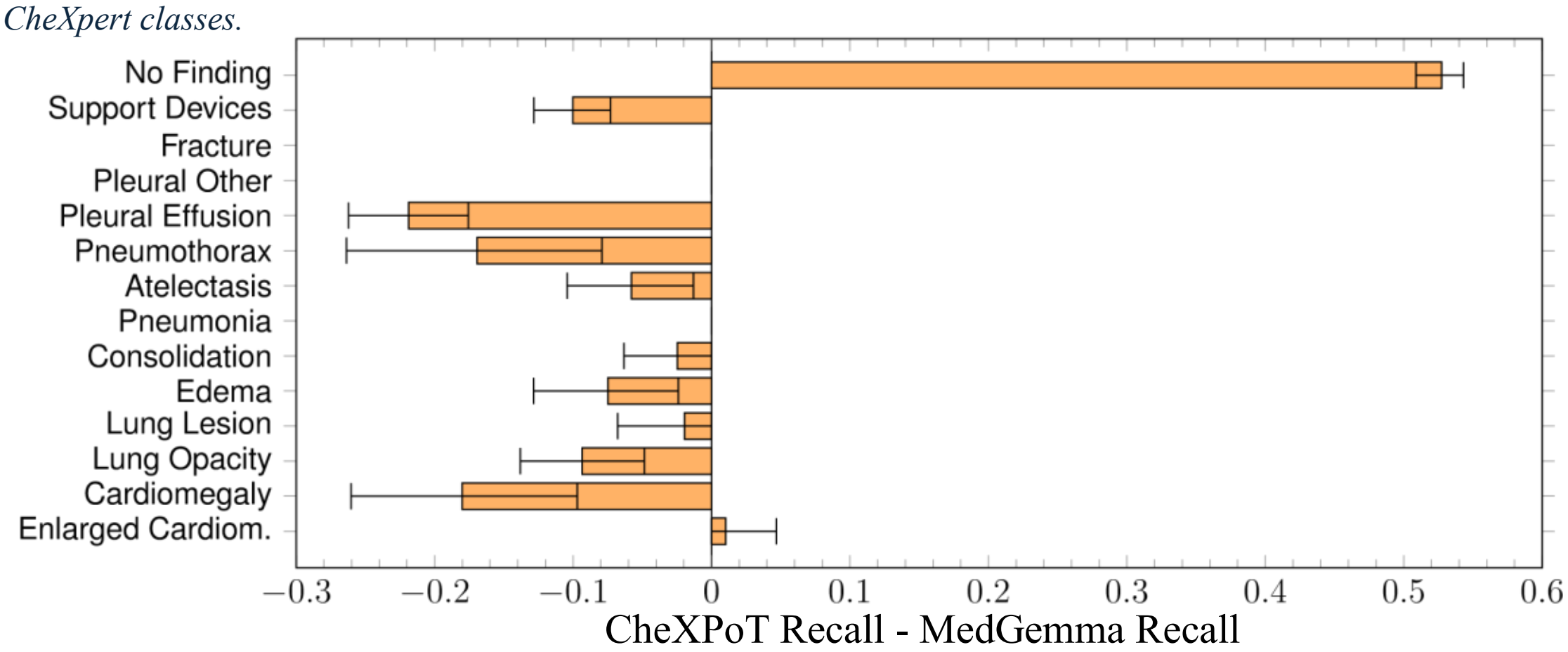

*Figure 5: 95% confidence intervals for difference in Recall between CheXPoT corrected reports and MedGemma Reports on the 14 CheXpert classes.*

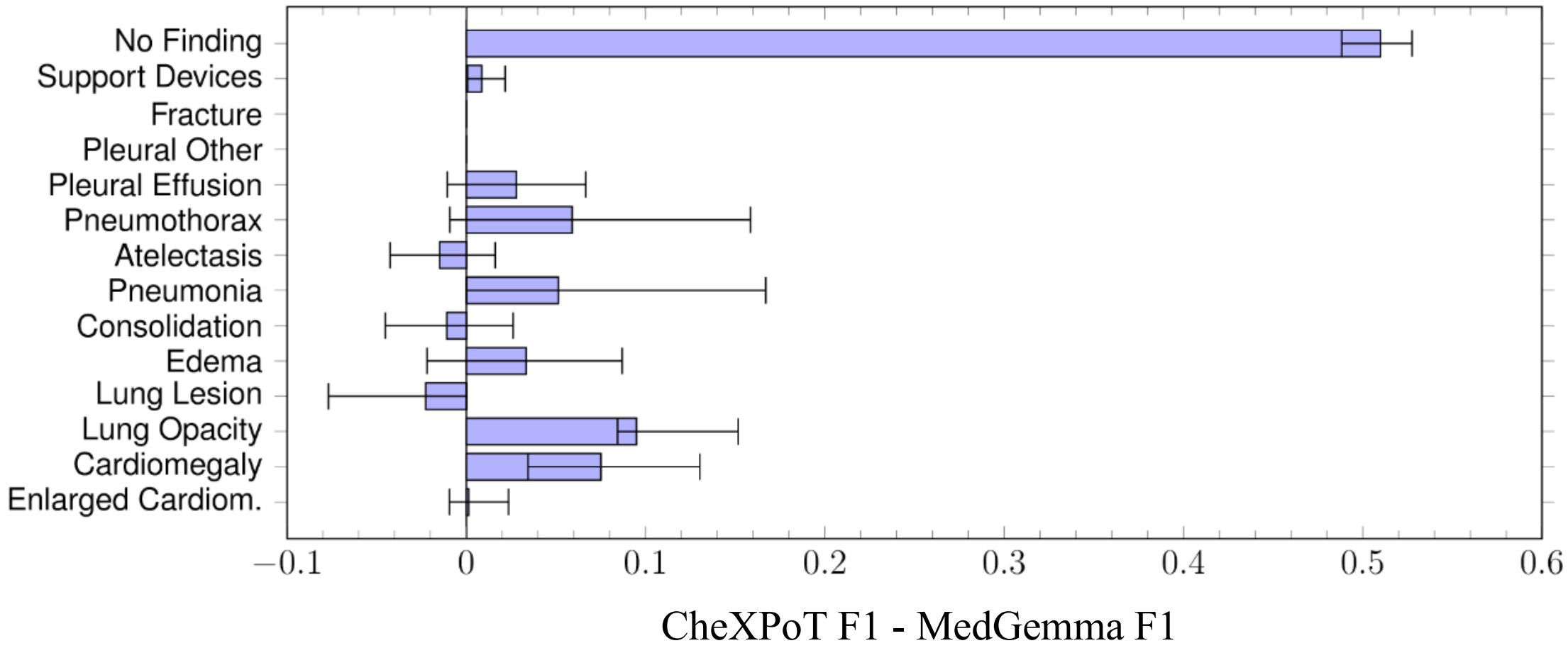

*Figure 6: 95% confidence intervals for difference in F1 Score between CheXPoT corrected reports and MedGemma Reports on the 14 CheXpert classes.*

## 3.1 Multi-Model Benchmarking

To assess whether CheXPoT generalizes beyond a single report generator, we benchmarked the framework on reports produced by MedGemma-4B, LLaVA [27], and MAIRA-2 [28] using the same 4,145-image held-out test split and evaluation protocol. For each model, we compared the original reports with their CheXPoT-corrected versions using CheXbert precision, recall, and F1 across the 14 CheXpert classes. This paired design isolates the effect of CheXPoT within each generator while also allowing comparison of how the correction framework performs across models with different baseline capabilities.

Figure 7 and Figure 8 show the summarized comparison of CheXPot Performance Improvement of Micro-Average Precision, Recall, and F1 Score on MedGemma-4B, LLaVA, and MAIRA-2.

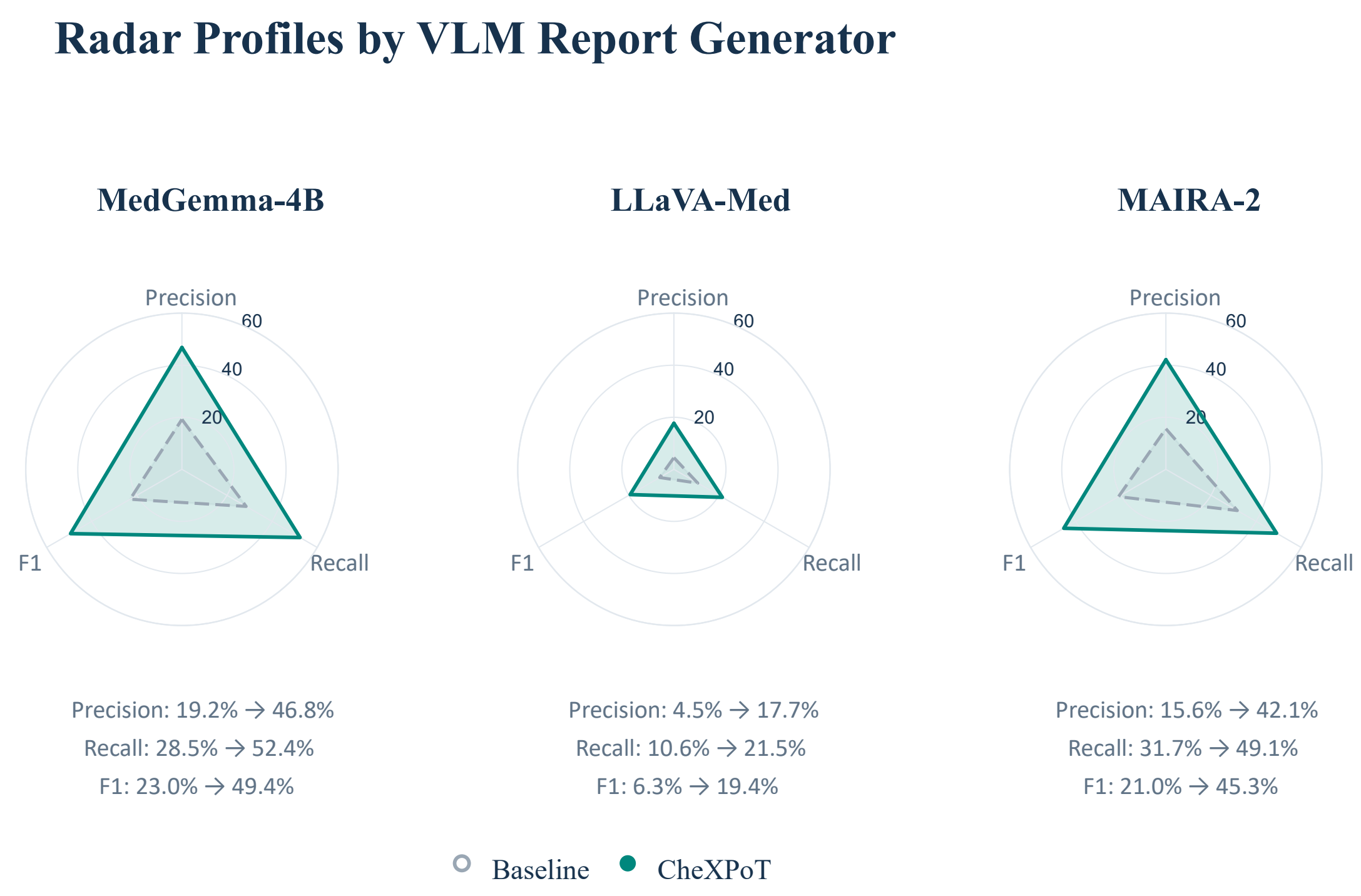

*Figure 7: Comparison of gains on micro-average Precision, recall and F1 score over 3 VLM report generators*

CheXPoT increased LLaVA micro-average precision from 0.045 to 0.177, micro-average recall from 0.106 to 0.215, and micro-average F1 from 0.063 to 0.194. CheXPoT significantly improved precision for six CheXpert classes: Enlarged Cardiomediastinum, Cardiomegaly, Edema, Atelectasis, Pneumothorax, and Pleural Effusion. It significantly improved recall for four classes: Enlarged Cardiomediastinum, Lung Opacity, Edema, and No Finding, and significantly improved F1 for six classes: Enlarged Cardiomediastinum, Cardiomegaly, Lung Opacity, Edema, Pleural Effusion, and No Finding. Recall decreased significantly for Pneumonia, Atelectasis, Pneumothorax, and Pleural Effusion; precision decreased significantly for No Finding. No statistically significant decreases were observed for F1.

CheXPoT increased MAIRA-2 micro-average precision from 0.156 to 0.421, micro-average recall from 0.317 to 0.491, and micro-average F1 from 0.210 to 0.453. CheXPoT significantly improved precision for seven CheXpert classes: Cardiomegaly, Lung Opacity, Edema, Atelectasis, Pneumothorax, Pleural Effusion, and No Finding. It significantly improved F1 for five classes: Cardiomegaly, Lung Opacity, Atelectasis, Pleural Effusion, and No Finding. No statistically significant decreases were observed for Precision or F1.

| Label | Support | CheXPoT + LLaVA | | | LLaVA | | | CheXPoT + MAIRA-2 | | | MAIRA-2 | | |
|---|---|---|---|---|---|---|---|---|---|---|---|---|---|
| | | Prec. | Rec. | F1 | Prec. | Rec. | F1 | Prec. | Rec. | F1 | Prec. | Rec. | F1 |
| Enlarged Cardiom. | 71 | **0.020*** | **0.099*** | **0.033*** | 0.000 | 0.000 | 0.000 | **0.033** | **0.028** | **0.030** | 0.014 | 0.014 | 0.014 |
| Cardiomegaly | 188 | **0.054*** | **0.633** | **0.100*** | 0.046 | 0.606 | 0.086 | **0.241*** | 0.383 | **0.296*** | 0.080 | 0.686 | 0.144 |
| Lung Opacity | 578 | **0.350** | **0.144*** | **0.204*** | 0.276 | 0.014 | 0.026 | **0.336*** | 0.246 | **0.284*** | 0.151 | 0.401 | 0.219 |
| Lung Lesion | 50 | **0.167** | **0.020** | **0.036** | 0.000 | 0.000 | 0.000 | **0.174** | 0.080 | **0.110** | 0.091 | 0.120 | 0.103 |
| Edema | 241 | **0.333*** | **0.037*** | **0.067*** | 0.000 | 0.000 | 0.000 | **0.320*** | 0.162 | **0.215** | 0.132 | 0.481 | 0.207 |
| Consolidation | 81 | 0.000 | 0.000 | 0.000 | 0.000 | 0.000 | 0.000 | **0.080** | 0.025 | 0.038 | 0.046 | 0.049 | 0.048 |
| Pneumonia | 13 | **0.009** | 0.154 | **0.017** | 0.002 | 0.538 | 0.004 | **0.000** | 0.000 | 0.000 | 0.000 | 0.000 | 0.000 |
| Atelectasis | 172 | **0.049*** | 0.448 | **0.089** | 0.043 | 0.558 | 0.079 | **0.083*** | 0.250 | **0.125*** | 0.052 | 0.500 | 0.094 |
| Pneumothorax | 101 | **0.231*** | 0.059 | **0.094** | 0.021 | 0.297 | 0.039 | **0.323*** | 0.099 | **0.152** | 0.108 | 0.248 | 0.151 |
| Pleural Effusion | 444 | **0.299*** | 0.417 | **0.348*** | 0.104 | 0.775 | 0.184 | **0.310*** | 0.376 | **0.340*** | 0.151 | 0.786 | 0.254 |
| Pleural Other | 17 | 0.000 | 0.000 | 0.000 | 0.000 | 0.000 | 0.000 | **0.150** | 0.176 | **0.162** | 0.118 | 0.235 | 0.157 |
| Fracture | 56 | **0.019** | 0.321 | **0.035** | 0.017 | 0.339 | 0.032 | **0.034** | 0.018 | **0.024** | 0.030 | 0.018 | 0.022 |
| Support Devices | 630 | 0.000 | 0.000 | 0.000 | 0.000 | 0.000 | 0.000 | **0.204** | 0.810 | 0.326 | 0.203 | 0.844 | 0.328 |
| No Finding | 3210 | 0.843 | **0.233*** | **0.365*** | 1.000 | 0.001 | 0.002 | **0.859*** | **0.584*** | **0.696*** | 0.818 | 0.116 | 0.203 |
| Micro avg. | 5852 | **0.177** | **0.215** | **0.194** | 0.045 | 0.106 | 0.063 | **0.421** | **0.491** | **0.453** | 0.156 | 0.317 | 0.210 |
| Macro avg. | 5852 | **0.170** | 0.183 | **0.099** | 0.108 | 0.224 | 0.032 | **0.225** | 0.231 | **0.200** | 0.142 | 0.321 | 0.139 |
| Weighted avg. | 5852 | 0.542 | **0.215** | **0.258** | 0.587 | 0.106 | 0.024 | **0.583** | **0.491** | **0.498** | 0.510 | 0.317 | 0.208 |
| Samples avg. | 5852 | **0.229** | **0.224** | **0.222** | 0.042 | 0.054 | 0.044 | **0.418** | **0.535** | **0.444** | 0.179 | 0.211 | 0.183 |

*Table 2: Per-label CheXbert evaluation of LLaVA-Med and MAIRA-2 reports before and after CheXPoT correction on the 4,145-image held-out test split. Green bold values indicate metrics where CheXPoT improves over the baseline VLM. Support is identical for both systems and denotes the number of positive label instances in the reference labels.*

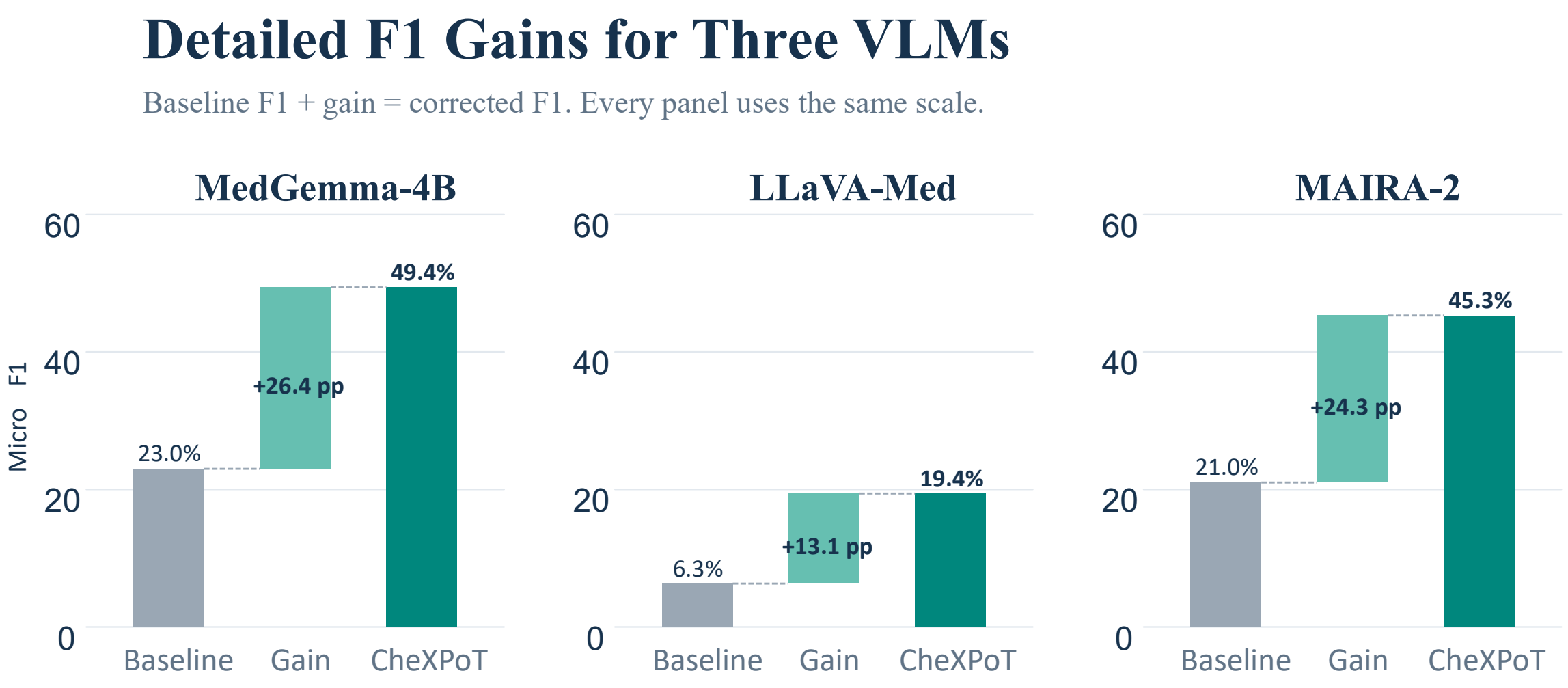

*Figure 8: Number-based gains on micro-average F1 score over 3 VLM report generators*

We also ran the same bootstrap tests, sampling from the test dataset with replacement 10,000 times and calculating the difference between CheXPoT's and LLaVA-Med's metrics and between CheXPoT's and MAIRA-2's metrics. We collected those results into 95% confidence interval plots, but for the ease of reading, those charts were moved to Appendix B.

Across the evaluated models, CheXPoT consistently improved report-level performance. The largest changes were observed for MedGemma-4b, where micro-average precision changed from 0.192 to 0.468 and micro-average F1 changed from 0.230 to 0.494. These results support the hypothesis that post-hoc visual verification can reduce unsupported findings across different VLMs. Variation of performance among models suggests that CheXPoT does not simply overwrite the report with the visual findings; CheXPoT does the correction based on the original report. In fact when closely examining the individual reports during debugging, we found that the corrected reports retain similar style and reporting format, differing based off of the unique VLM generator.

# 4 Discussion

We present CheXPoT as an early post-hoc verification layer for MedGemma-4B radiology report generation. CheXPoT operates after the base VLM has generated a report, checking whether generated findings are visually supported before producing a revised report.

CheXPoT produced clearer gains in visual factuality. It significantly improved precision for Cardiomegaly, Lung Opacity, Edema, Pneumonia, Pneumothorax, Pleural Effusion, Support Devices, and No Finding. It also significantly improved F1 for Cardiomegaly, Lung Opacity, Pneumonia, Support Devices, and No Finding. These labels include both common findings and clinically salient abnormalities, suggesting that visual verification may be most useful when unsupported positive claims would otherwise make the report appear more complete than the image evidence supports. The smaller number of F1 improvements compared with precision improvements is consistent with a precision–recall tradeoff: stricter verification can remove hallucinated findings, but it may also remove some true positives.

To evaluate whether mapping free-text report predicates to a structured 14-label evidence table introduces coverage gaps, we conducted a comprehensive empirical validation. We analyzed all 4,145 raw clinical reports generated by MedGemma using RadGraph to extract all clinical entities and relations. Out of these 4,145 reports, only 48 mentions of Chronic Lung Disease and 8 mentions of Scoliosis fell outside the standard 14 CheXpert Plus categories. For the 8 instances of Scoliosis reporting, CheXpert rightfully omitted those because it is not cardiopulmonary pathology. For the 48 mentions of Chronic Lung Disease, 43 were mentioned together with Edema or other official CheXpert Pathologies, e.g., *"... possibly representing pulmonary edema or chronic lung disease ... ,"* meaning those 43 are not ignored, thus only 5 reports of chronic lung disease were removed. This demonstrates that the 14-label schema captures virtually all useful clinical findings from the original MedGemma outputs, which has trivial impact on the final significance of our results.

CheXPoT provides a mechanism for correcting unsupported claims rather than only flagging them. These results should be interpreted as evidence that post-hoc visual verification is feasible for MedGemma-4B outputs.

# 5 Limitations

This study has several limitations that also define the most important next steps. First, all experiments use a single primary dataset, CheXpert Plus. The results may not generalize to other radiology datasets or different institutional reporting styles.

Second, CheXPoT is reliant on the code created by the LLM. During this experiment, no code sandboxing was performed, so a malicious piece of code such as `shutil.rmtree(...)` could potentially cause catastrophic consequences. Further implementations of CheXPoT should implement code sandboxing to avoid these potential risks. The code generated by the LLM could also potentially raise an error. Further implementations of CheXPoT should implement exception handling to avoid these potential risks. Neither of these two types of incidents were observed during our experiment.

Third, CheXPoT is evaluated through CheXbert label extraction rather than direct clinical review. CheXbert enables scalable measurement across the 14 CheXpert labels, but it cannot fully assess report usefulness, severity, temporal comparisons, phrasing quality, or clinically important details outside those labels. CheXPoT is also limited by the DenseNet verifier, so errors in the visual evidence table can propagate into the rewrite.

Finally, CheXPoT has not been tested in a clinical workflow. A blind radiologist review would be needed to assess whether visually verified rewrites improve trust, efficiency, or diagnostic safety. Until such evaluation is performed, CheXPoT should be viewed as a research prototype for post-hoc trust infrastructure.

# 6 Conclusion

We introduce CheXPoT, a novel post-hoc hallucination-correction framework for medical VLMs that validates generated radiology findings against image-derived evidence. Through our innovative method of checking and correcting unsupported claims from MedGemma-4B, LLaVA-Med, and MAIRA-2 chest X-ray reports, CheXPoT significantly improves precision and F1 across multiple CheXpert labels for all three base VLMs, suggesting that lightweight visual verification can make generated reports more reliable without retraining the base model. More broadly, CheXPoT lays a foundation for safer and more trustworthy tools that help validate VLM-generated radiology reports, reduce unsupported findings, and support radiologists in model-assisted workflows. Future work should evaluate CheXPoT across additional models, datasets, and prospective clinical settings to determine whether these gains translate into reduced physician burden and improved patient care.

# Acknowledgements

We acknowledge the Stanford AIMI Center for supporting this research and providing the computer infrastructure to execute experiments. I also thank my mentors Dr. Alaa Youssef and Rayan Ansari for their invaluable directional feedback throughout this research.

# References

[1] S. Zhang *et al.*, "RadFlag: A black-box hallucination detection method for medical vision language models," in *Proc. 4th Machine Learning for Health Symp. (ML4H)*, 2025, arXiv:2411.00299. [Online]. Available: https://arxiv.org/abs/2411.00299

[2] R. Yan *et al.*, "Worse than random? An embarrassingly simple probing evaluation of large multimodal models in medical VQA," 2024, arXiv:2406.10185. [Online]. Available: https://arxiv.org/abs/2406.10185

[3] "Hallucination rate quantification in radiology AI," *npj Digital Medicine*, 2025. [Online]. Available: https://www.nature.com/articles/s41746-025-01670-7

[4] Z. Chen *et al.*, "CheXagent: Towards a foundation model for chest X-ray interpretation," 2024, arXiv:2401.12208. [Online]. Available: https://arxiv.org/abs/2401.12208

[5] S. Bannur *et al.*, "MAIRA-2: Grounded radiology report generation," 2024, arXiv:2406.04449. [Online]. Available: https://arxiv.org/abs/2406.04449

[6] R. Zhao, R. Xia, and P. Li, "Hallucination mitigating for medical report generation," 2026, arXiv:2601.15745. [Online]. Available: https://arxiv.org/abs/2601.15745

[7] O. Banerjee *et al.*, "Direct preference optimization for suppressing hallucinated prior exams in radiology report generation," in *Proc. 9th Machine Learning for Healthcare Conf. (MLHC)*, 2024, arXiv:2406.06496. [Online]. Available: https://arxiv.org/abs/2406.06496

[8] Z. Li *et al.*, "The hidden life of tokens: Reducing hallucination of large vision-language models via visual information steering," in *Proc. ICML*, 2025, arXiv:2502.03628. [Online]. Available: https://arxiv.org/abs/2502.03628

[9] R. Hardy, S. E. Kim, D. H. Ro, and P. Rajpurkar, "ReXTrust: A model for fine-grained hallucination detection in AI-generated radiology reports," in *Proc. First AAAI Bridge Program on AI for Medicine and Healthcare*, 2025, arXiv:2412.15264. [Online]. Available: https://arxiv.org/abs/2412.15264

[10] P. Wienholt *et al.*, "Hallucination filtering in radiology vision-language models using discrete semantic entropy," *Eur. Radiol.*, 2026, arXiv:2510.09256. [Online]. Available: https://arxiv.org/abs/2510.09256
[11] H. Wang, Y. Zhou, B. Ghosh, K. Batmanghelich, and X. Li, "Semantic consistency-based uncertainty quantification for factuality in radiology report generation," in *Findings of ACL: NAACL*, 2025, arXiv:2412.04606. [Online]. Available: https://arxiv.org/abs/2412.04606
[12] J. Pan *et al.*, "MedVLM-R1: Incentivizing medical reasoning capability of vision-language models (VLMs) via reinforcement learning," 2025, arXiv:2502.19634. [Online]. Available: https://arxiv.org/abs/2502.19634
[13] NVIDIA, "NV-Reason-CXR-3B: A reasoning visual language model for chest X-ray analysis," 2025, arXiv:2510.23968. [Online]. Available: https://arxiv.org/abs/2510.23968
[14] Q. Gu *et al.*, "ChestX-Reasoner: Advancing radiology foundation models with reasoning through step-by-step verification," 2025, arXiv:2504.20930. [Online]. Available: https://arxiv.org/abs/2504.20930
[15] L. Yan *et al.*, "X-Ray-CoT: Interpretable chest X-ray diagnosis with vision-language models via chain-of-thought reasoning," 2025, arXiv:2508.12455. [Online]. Available: https://arxiv.org/abs/2508.12455
[16] R. Rafailov *et al.*, "Direct preference optimization: Your language model is secretly a reward model," in *Proc. NeurIPS*, 2023, arXiv:2305.18290. [Online]. Available: https://arxiv.org/abs/2305.18290
[17] P. Sahu, K. Sikka, and A. Divakaran, "Pelican: Correcting hallucination in vision-LLMs via claim decomposition and program-of-thought verification," 2024, arXiv:2407.02352. [Online]. Available: https://arxiv.org/abs/2407.02352
[18] S. Jain *et al.*, "RadGraph: Extracting clinical entities and relations from radiology reports," in *Proc. NeurIPS Datasets and Benchmarks Track*, 2021. [Online]. Available: https://arxiv.org/abs/2106.14463
[19] G. Huang, Z. Liu, L. van der Maaten, and K. Q. Weinberger, "Densely connected convolutional networks," in *Proc. IEEE Conf. Computer Vision and Pattern Recognition (CVPR)*, Jul. 2017, pp. 4700–4708.
[20] R. R. Selvaraju, M. Cogswell, A. Das, R. Vedantam, D. Parikh, and D. Batra, "Grad-CAM: Visual explanations from deep networks via gradient-based localization," in *Proc. IEEE Int. Conf. Computer Vision (ICCV)*, Oct. 2017, pp. 618–626.
[21] P. Chambon *et al.*, "CheXpert Plus: Augmenting a large chest X-ray dataset with text radiology reports, patient demographics and additional image formats," 2024, arXiv:2405.19538. [Online]. Available: https://arxiv.org/abs/2405.19538
[22] Stanford Center for Artificial Intelligence in Medicine & Imaging, "CheXpert Plus," 2024. [Online]. Available: https://aimi.stanford.edu/datasets/chexpert-plus
[23] A. Smit, S. Jain, P. Rajpurkar, A. Pareek, A. Y. Ng, and M. P. Lungren, "CheXbert: Combining automatic labelers and expert annotations for accurate radiology report labeling using BERT," 2020, arXiv:2004.09167. [Online]. Available: https://arxiv.org/abs/2004.09167
[24] T.-Y. Lin, P. Goyal, R. Girshick, K. He, and P. Dollar,´ "Focal loss for dense object detection," in *Proc. IEEE Int. Conf. Computer Vision (ICCV)*, 2017, arXiv:1708.02002. [Online]. Available: https://arxiv.org/abs/1708.02002
[25] I. Loshchilov and F. Hutter, "Decoupled weight decay regularization," in *Proc. ICLR*, 2019, arXiv:1711.05101. [Online]. Available: https://arxiv.org/abs/1711.05101
[26] Google Health, "MedGemma," GitHub repository, 2024. [Online]. Available: https://github.com/Google-Health/medgemma
[27] C. Li *et al.*, "LLaVA-Med: Training a Large Language-and-Vision Assistant for Biomedicine in One Day," 2023, arXiv:2306.00890. [Online]. Available: https://arxiv.org/abs/2306.00890
[28] S. Bannur *et al.*, "MAIRA-2: Grounded Radiology Report Generation," 2024, arXiv:2406.04449. [Online]. Available: https://arxiv.org/abs/2406.04449

# Appendix A: Example CheXPoT Pipeline Run

This appendix shows one representative CheXPoT execution trace, including predicate extraction, the visual evidence table, generated Program-of-Thought (PoT) code, and the resulting verification answers.

## A.1 Claim Decomposition Predicates

- **Devices**
  - `Count(surgical_device)`
  - `Exists(surgical_device)`
  - `Position(surgical_device, chest)`

- **Chest tube**
  - `Exists(chest_tube)`
  - `Attribute(chest_tube, isRightSided)`
  - `Position(chest_tube, chest)`
- **Pleural effusion**
  - `Exists(pleural_effusion)`
  - `Attribute(pleural effusion, isProminent)`
  - `Position(pleural_effusion, pleural)`
- **Underlying lung disease**
  - `Exists(underlying_lung_disease)`
  - `Position(underlying_lung_disease, lung)`
- **Pneumothorax**
  - `Exists(pneumothorax) == false`
- **Cardiac silhouette**
  - `Exists(cardiac_silhouette)`
  - `Attribute(cardiac_silhouette, isUnremarkable)`
- **Mediastinum**
  - `Exists(mediastinum)`
  - `Attribute(mediastinum, isUnremarkable)`
- **Other abnormalities**
  - `Exists(other_abnormalities) == false`

## A.2 Visual Evidence Table

*Table 4: Example visual table passed to PoT verification. Coordinates denote representative Grad-CAM points.*

| Finding | Present | Representative point |
|---|---|---|
| Enlarged Cardiomediastinum | 1 | (0.122, 0.122) |
| Cardiomegaly | 0 | (0.122, 0.122) |
| Lung Opacity | 0 | (0.000, 0.000) |
| Lung Lesion | 1 | (0.958, 0.122) |
| Edema | 0 | (0.122, 0.122) |
| Consolidation | 1 | (0.122, 0.000) |
| Pneumonia | 0 | (0.122, 0.122) |
| Atelectasis | 1 | (0.122, 0.122) |
| Pneumothorax | 0 | (0.958, 0.000) |
| Pleural Effusion | 0 | (0.208, 0.375) |
| Pleural Other | 1 | (0.958, 0.122) |
| Fracture | 1 | (0.122, 0.122) |
| Support Devices | 0 | (0.122, 0.122) |
| No Finding | 0 | (0.122, 0.122) |

## A.3 Generated PoT Code

```
output_s = {}

# Devices: There is no "surgical_device" in the allowed names, but "Support Devices" is present.
# We'll use "Support Devices" as the closest match.
output_s['Count(surgical_device)'] = count_abnormality("Support
Devices") output_s['Exists(surgical_device)'] =
exists_abnormality("Support Devices") # Position relative to "chest" is
not answerable (no "chest" object), so skip.

# Chest Tube: Not present in allowed names, so skip all chest tube questions.
output_s['Exists(chest_tube)'] = "no answer was given for this question"
output_s['Attribute(chest_tube, isRightSided)'] = "no answer was given for this question"
output_s['Position(chest_tube, chest)'] = "no answer was given for this question"
```

```
# Pleural Effusion: output_s['Exists(pleural_effusion)'] =
exists_abnormality("Pleural Effusion") # "isProminent" attribute is not
directly answerable, so skip.
output_s['Attribute(pleural_effusion, isProminent)'] = "no answer was given for this
question" # "pleural" is not an object, so skip position.
output_s['Position(pleural_effusion, pleural)'] = "no answer was given for this
question"

# Underlying Lung Disease: Not present in allowed names, but we can check for any lung
abnormality.
lung_abnormalities = [
    "Lung Opacity", "Lung Lesion", "Edema", "Consolidation", "Pneumonia", "Atelectasis"
] exists_lung_disease = any(exists_abnormality(name) for name in lung_abnormalities)
output_s['Exists(underlying_lung_disease)'] = exists_lung_disease # "lung" is not an
object, so skip position. output_s['Position(underlying_lung_disease, lung)'] = "no
answer was given for this question"

# Pneumothorax: output_s['Exists(pneumothorax) == false'] = not
exists_abnormality("Pneumothorax")

# Cardiac Silhouette: Not present in allowed names, but "Cardiomegaly" and "Enlarged
    Cardiomediastinum" are related.
# We'll use their absence to infer "Exists(cardiac_silhouette)" and "isUnremarkable".
exists_cardiac = exists_abnormality("Cardiomegaly") or exists_abnormality("Enlarged
Cardiomediastinum") output_s['Exists(cardiac_silhouette)'] = True # The silhouette is always
present in a chest x-ray.
output_s['Attribute(cardiac_silhouette, isUnremarkable)'] = not exists_cardiac

# Mediastinum: Not present in allowed names, but "Enlarged Cardiomediastinum" is related.
exists_mediastinum = exists_abnormality("Enlarged Cardiomediastinum")
output_s['Exists(mediastinum)'] = True # The mediastinum is always present.
output_s['Attribute(mediastinum, isUnremarkable)'] = not exists_mediastinum

# Other Abnormalities: Not present in allowed names, but we can check if any abnormality except
     "No Finding" and "Support Devices" is present.
other_abnormalities = [
    "Enlarged Cardiomediastinum", "Cardiomegaly", "Lung Opacity", "Lung Lesion", "Edema",
    "Consolidation", "Pneumonia", "Atelectasis", "Pneumothorax", "Pleural Effusion",
    "Pleural Other", "Fracture"
] exists_other_abnormalities = any(exists_abnormality(name) for name in other_abnormalities)
output_s['Exists(other_abnormalities) == false'] = not exists_other_abnormalities output_s
```

### A.4 PoT Verification Answers

```
{'Count(surgical_device)': 0, 'Exists(surgical_device)': False, 'Exists(pleural_effusion)':
     False, 'Exists( underlying_lung_disease)': True, 'Exists(pneumothorax) == false':
     True, 'Exists(cardiac_silhouette)': True, 'Attribute(cardiac_silhouette, isUnremarkable)':
     False, 'Exists(mediastinum)': True, 'Attribute( mediastinum, isUnremarkable)':
     False, 'Exists(other_abnormalities) == false': False}
```

# Appendix B: 95% Confidence Intervals for Difference in Performance for Other Report Generators

## B.1 LLaVA-Med

Figures 9 and show the paired bootstrap gains and 95% confidence intervals for CheXPoT-corrected reports relative to the uncorrected LLaVA reports.

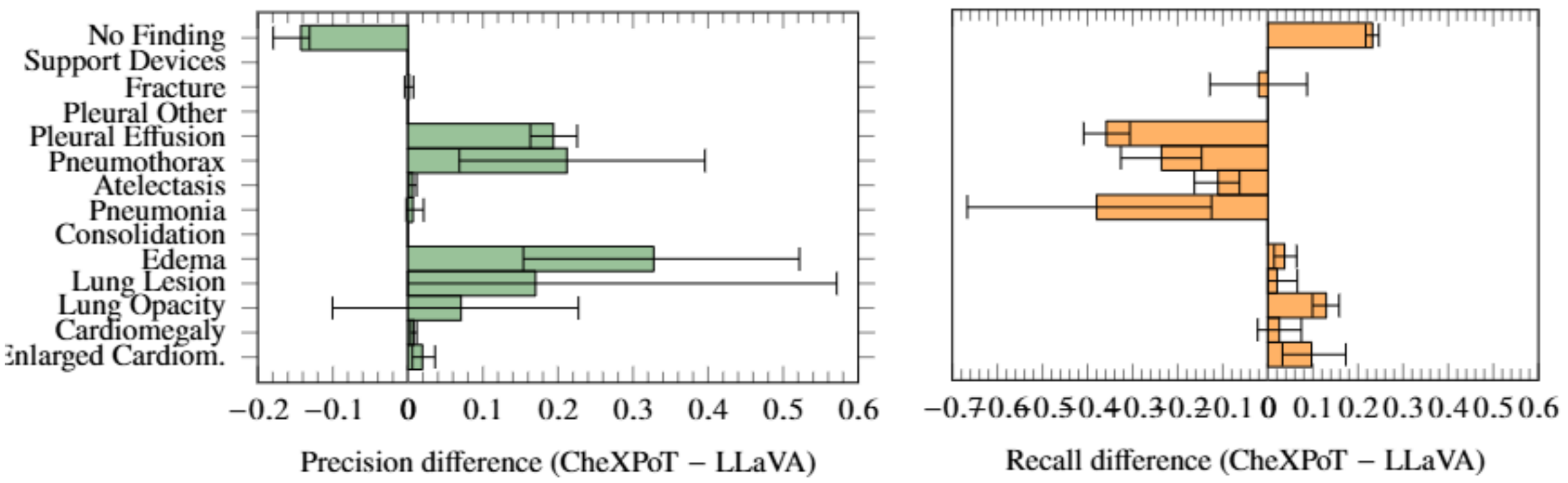

*Figure 9: 95% confidence intervals for differences in precision (left) and recall (right) between CheXPoT-corrected and uncorrected LLaVA reports across the 14 CheXpert classes.*

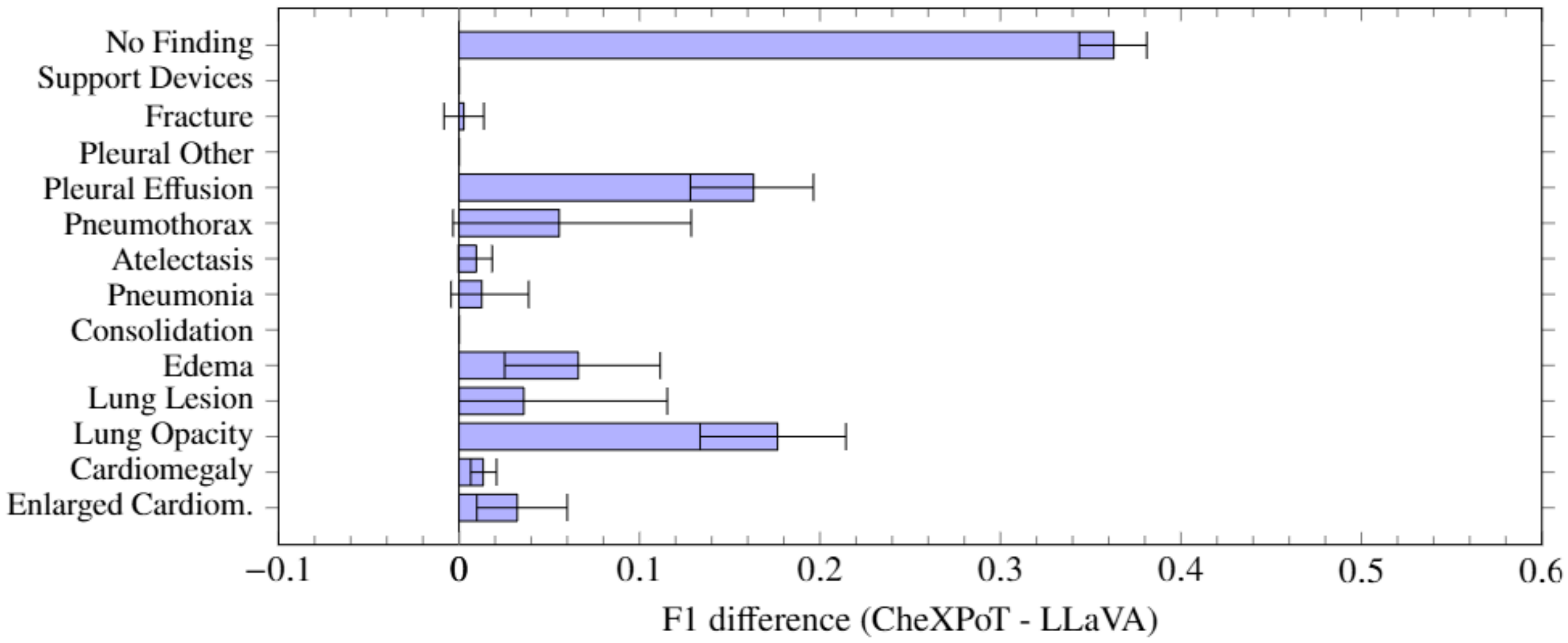

*Figure 10: 95% confidence intervals for F1 differences between CheXPoT-corrected and uncorrected LLaVA reports across the 14 CheXpert classes.*

## B.2 MAIRA-2

Figures 11 and 12 show the paired bootstrap gains and 95% confidence intervals for CheXPoT-corrected reports relative to the uncorrected MAIRA-2 reports.

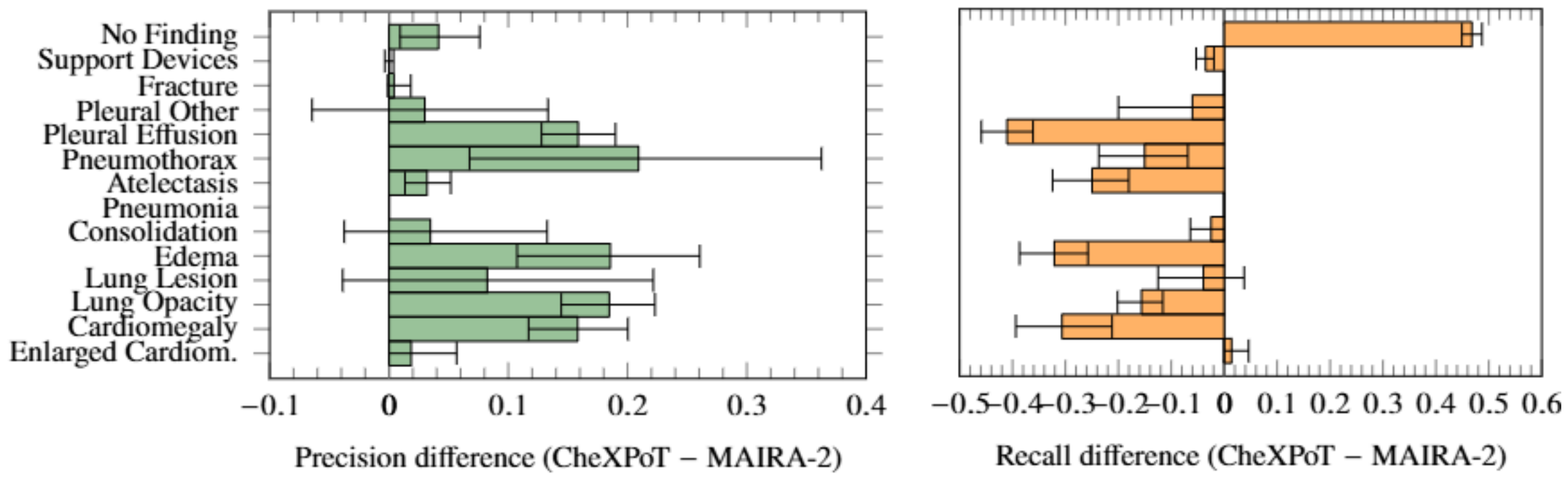

*Figure 11: 95% confidence intervals for differences in precision (left) and recall (right) between CheXPoT-corrected and uncorrected MAIRA-2 reports across the 14 CheXpert classes.*

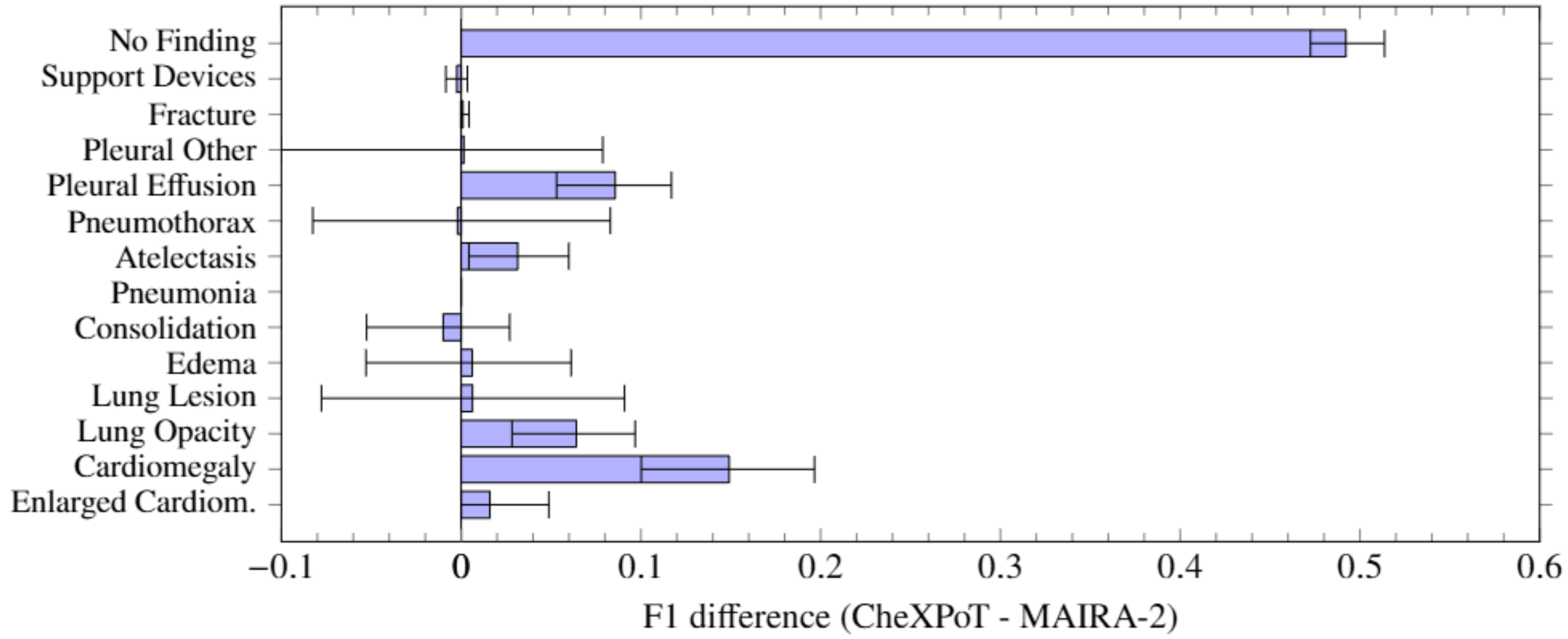

*Figure 12: 95% confidence intervals for F1 differences between CheXPoT-corrected and uncorrected MAIRA-2 reports across the 14 CheXpert classes.*